\pdfoutput=1

\documentclass[11pt]{article}

\usepackage[final]{acl}

\usepackage{times}
\usepackage{latexsym}

\usepackage[T1]{fontenc}

\usepackage[utf8]{inputenc}

\usepackage{microtype}

\usepackage{inconsolata}

\usepackage{graphicx}

\title{\textit{TALL}: A Trainable Architecture for Enhancing LLM Performance\\ in Low-Resource Languages}

\author{
  \textbf{Moshe Ofer\textsuperscript{}},
  \textbf{Orel Zamler\textsuperscript{}},
  \textbf{Amos Azaria\textsuperscript{}}
\\
\\
  \textsuperscript{}Ariel University
\\
  \small{
    \textbf{Correspondence:} \href{mailto:mosheo@ariel.ac.il}{mosheo@ariel.ac.il}, \href{mailto:amos.azaria@ariel.ac.il}{amos.azaria@ariel.ac.il}
  }
}

\begin{document}
\maketitle
\begin{abstract}
Large Language Models (LLMs) excel in high-resource languages but struggle with low-resource languages due to limited training data. This paper presents \textit{TALL} (\textbf{T}rainable \textbf{A}rchitecture for Enhancing \textbf{L}LM Performance in \textbf{L}ow-Resource Languages), which integrates an LLM with two bilingual translation models. TALL transforms low-resource inputs into high-resource representations, leveraging the LLM's capabilities while preserving linguistic features through dimension alignment layers and custom transformers. Our experiments on Hebrew demonstrate significant improvements over several baselines, including direct use, naive translation, and fine-tuning approaches. The architecture employs a parameter-efficient strategy, freezing pre-trained components while training only lightweight adapter modules, balancing computational efficiency with performance gains.\footnote{Our code is available at \url{https://github.com/MosheOfer1/TALL}.}
\end{abstract}

\section{Introduction}
\label{sec:introduction}

Large Language Models (LLMs) have significantly advanced natural language processing (NLP) for high-resource languages like English, but their performance in low-resource languages remains limited due to sparse data, linguistic complexity, and the scarcity of annotated datasets and pre-trained models. Efforts to address these challenges have primarily focused on transfer learning \cite{5288526}, which leverages knowledge from high-resource languages to enhance performance in low-resource settings, though notable limitations persist due to linguistic diversity and persistent data scarcity.

This paper introduces \textit{TALL}, a novel architecture that harnesses the strengths of LLMs trained on high-resource languages to improve performance on low-resource languages. The key innovation lies in integrating three pre-trained models: an LLM trained primarily on a high-resource language, and two translation models that bridge the gap between high and low-resource languages. Through carefully designed dimension alignment adapters \cite{DBLP:journals/corr/RebuffiBV17, DBLP:journals/corr/abs-1902-00751} and custom transformer \cite{DBLP:journals/corr/VaswaniSPUJGKP17} components, \textit{TALL} ensures that the linguistic knowledge embedded in high-resource LLMs benefits low-resource language processing.

We validate \textit{TALL} through experiments on Hebrew, chosen for its rich morphology, complex syntax, and limited available datasets. Our results demonstrate significant improvements in accuracy compared to direct use of LLMs, naive translation approaches, soft prompting, and fine-tuning, highlighting the effectiveness of our architecture.

\label{sec:related_work}

\section{Related Work}

Our work builds upon several approaches to improving LLM performance in low-resource settings:

\paragraph{Cross-lingual Transfer} Prior research has explored transferring knowledge from high-resource to low-resource languages through various methods. Cross-lingual in-context learning \cite{cahyawijaya2024llms} leverages examples from high-resource languages to improve performance in low-resource contexts. However, these approaches often lack the structural integration that \textit{TALL} provides.

\paragraph{Parameter-Efficient Adaptation} Adapter-based approaches \cite{DBLP:journals/corr/abs-2005-00247, pfeiffer-etal-2020-mad} insert small, trainable modules into frozen pre-trained models to adapt them to new tasks or languages. Our dimension alignment adapters follow this principle, but specifically focus on aligning representations between different language spaces.

\paragraph{Translation-based Strategies} Recent work has introduced creative translation-based strategies. Notable examples include TaCo~\cite{upadhayay2024taco}, which integrates translation into reasoning; CoTR~\cite{cotr}, which translates inputs into a high-resource language for processing; and TALENT~\cite{guo-etal-2024-teaching}, which employs a curriculum-inspired approach to grammar learning before translation.

\paragraph{Concept-level Processing} Large Concept Models \cite{the2024large} operate on sentence-level "concepts" in a unified multilingual embedding space, offering an alternative to token-level prediction. Similar to \textit{TALL}, they aim to create cross-lingual representations that preserve semantic meaning across languages. 

\section{Model Architecture}
\label{sec:model_architecture}

\subsection{Architecture Overview}
\label{sec:architecture_overview}

\textit{TALL} strategically integrates pre-trained components with lightweight, trainable adapters to efficiently process low-resource language inputs. As illustrated in Figure~\ref{fig:tall_architecture}, the architecture consists of seven main stages (numbered 1-7 in the figure):

\begin{figure}[t!]
    \centering
    \includegraphics[width=\columnwidth]{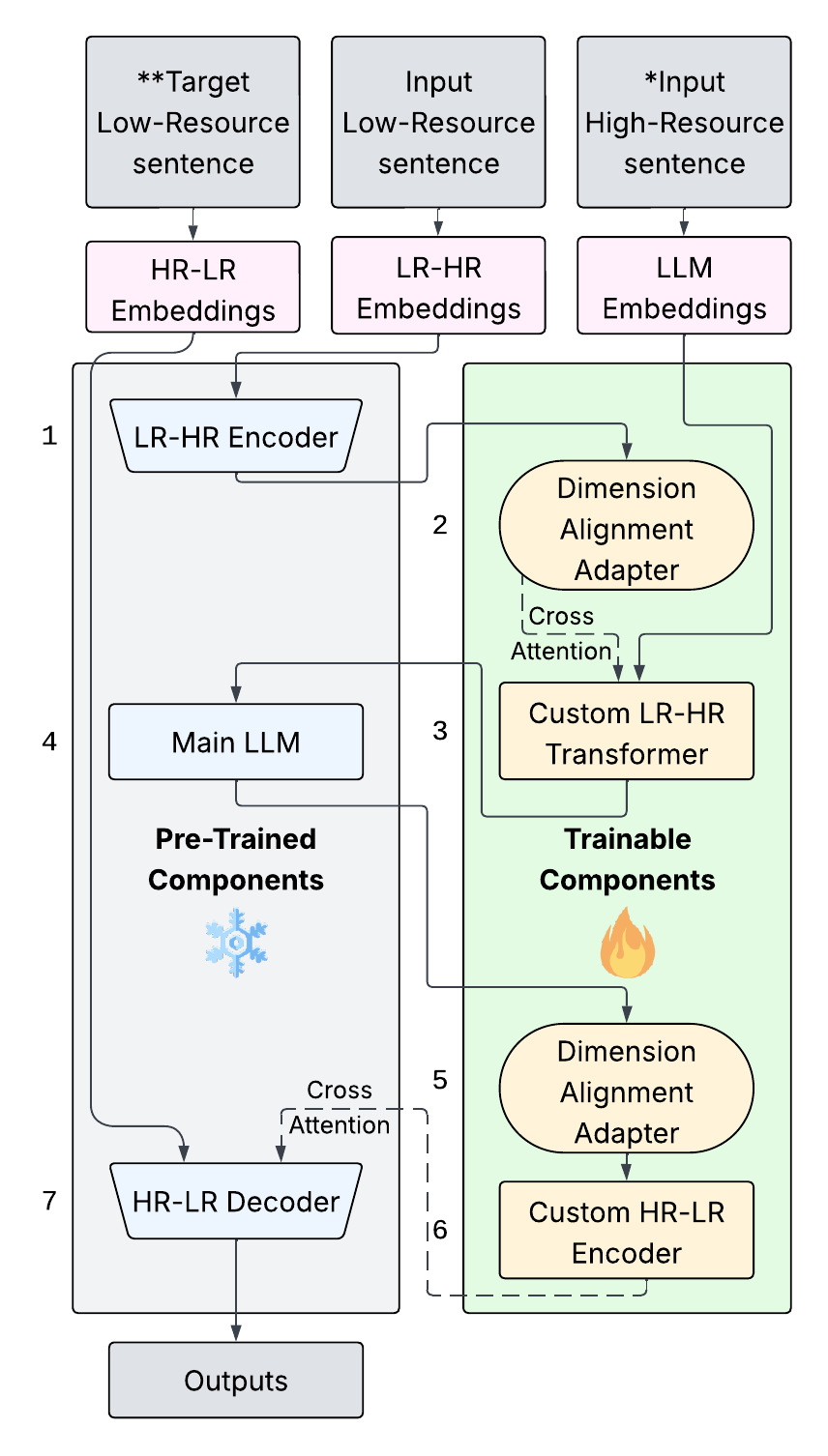}
    \caption{Overview of the TALL architecture with numbered components (1-7) corresponding to the processing stages described in Section \ref{sec:architecture_overview}. 
    * For input high-resource sentences, the last word is removed before translation during training; 
    ** For target sentences, teacher forcing is used during training while outputs are generated auto-regressively during inference.}
    \label{fig:tall_architecture}
\end{figure}

\begin{enumerate}
    \item \textbf{Source Language Encoding:} The input sentence in the low-resource language is processed by a pre-trained low-to-high-resource (LR-HR) encoder, generating hidden states that represent the intermediate semantic form.
    
    \item \textbf{Dimension Alignment (Source to LLM):} A trainable dimension alignment adapter transforms the encoder's hidden states to match the dimensional requirements of the LLM, ensuring seamless integration.
    
    \item \textbf{Custom Transformer (LR-HR):} This trainable component receives the LLM tokenized translated sentence embeddings (during training, the last word is removed before the sentence is translated) and applies cross attention \cite{DBLP:journals/corr/abs-2104-08771} to the aligned hidden states, ensuring the LLM receives representations enriched with information from the original input.
    
    \item \textbf{LLM Integration:} The processed sequence is passed to the frozen LLM for next-token prediction, leveraging its strong language modeling capabilities in the high-resource language.
    
    \item \textbf{Dimension Alignment (LLM to Target):} A second alignment adapter transforms the LLM's hidden states to match the dimensional requirements of the second translator.
    
    \item \textbf{Custom Transformer (HR-LR):} This component processes the aligned hidden states from the LLM, preparing them for the final decoding stage.
    
    \item \textbf{Target Language Decoding:} The HR-LR decoder generates the final text in the target low-resource language by applying cross-attention between the encoded representation and the target language embeddings.
\end{enumerate}

The implementation details for each component are as follows:

\paragraph{Translation Models} We use pre-trained Marian MT models \cite{junczys2018marian} for translation between Hebrew and English, leveraging their robust bilingual capabilities.

\paragraph{Dimension Alignment Adapters} These adapters consist of trainable fully-connected layers with layer normalization and GELU activations. They project hidden states between the different dimensional spaces required by each pre-trained component.

\paragraph{Custom Transformers} We implement two custom transformer modules derived from Marian configurations: (1) a transformer that conditions on high-resource tokens and cross-attends to dimension-aligned low-resource encoder states, and (2) a transformer that refines representations before the final decoding stage.

\paragraph{Parameter Efficiency} A key advantage of \textit{TALL} is its parameter efficiency. By freezing the large pre-trained components (translation models and LLM) and training only the lightweight alignment adapters and custom transformers, we minimize computational overhead while maximizing performance gains. In our implementation with bloomz-560m \cite{muennighoff2022crosslingual}, only 14.35\% of the total parameters (126.8M out of 883.5M) are trainable, while for QWEN2.5-0.5b \cite{qwen2.5}, 13.47\% of the total parameters (107.7M out of 799.2M) are trainable.

\begin{table*}[!t]
\centering
\begin{tabular}{llccc}
\hline
\textbf{Dataset} & \textbf{Approach} & \textbf{Bloomz-560m} & \textbf{QWEN2.5-0.5b} \\
\hline
WIKI-SLVM
    & Direct Hebrew     & 0.63  & 3.77  \\
    & Fine-tuned        & 0.16    & 2.75  \\
    & Naive             & 2.11  & 2.85  \\
    & Soft prompt       & 2.50  & 1.96    \\
    & From Scratch     & 2.93  & 3.99  \\
    & \textit{TALL}     & \textbf{5.59}  & \textbf{5.15} \\
\hline
HE-LYRICS
    & Direct Hebrew     & 0.14  & 2.11  \\
    & Fine-tuned        & 0.36    & 2.22  \\
    & Naive             & 1.83  & 2.20  \\
    & Soft prompt       & 2.05  & 1.90    \\
    & From Scratch     & 2.08  & 2.92  \\
    & \textit{TALL}     & \textbf{5.44}  & \textbf{4.77} \\
\hline
\end{tabular}
\caption{Accuracy comparison (\%) for predicting the missing Hebrew word. TALL, soft prompt, fine-tuning, and from-scratch were trained on the news dataset.
}
\label{tab:approach_accuracy_comparison}
\end{table*}
A detailed breakdown of the TALL parameters, including overall statistics and module-wise counts for both bloomz TALL and Qwen TALL, is provided in Tables~\ref{tab:overall_stats}, \ref{tab:bloomz_breakdown}, and \ref{tab:qwen_breakdown} in the Appendix.

\subsection{Training Procedure}

We trained \textit{TALL} on 256,000 Hebrew sentences from news articles. The dataset underwent standard preprocessing, including removal of duplicates and filtering for sentences between 5-30 words in length.

The model is trained using teacher forcing, where ground-truth target tokens are provided as decoder inputs. Importantly, the loss is computed only for the final missing token in the sequence. This limitation is inherent to the TALL architecture design: we translate the truncated sentence (without the final word) into the high-resource language and feed it to the Custom Transformer (LR-HR). Since the rest of the tokens are already presented as the high-resource language translation, we can only compute the prediction signal for the missing final token. While it would be preferable to obtain training signals from all tokens (as in standard language modeling), the current cross-lingual architecture constrains us to optimizing for the final token prediction only (see Section \ref{sec:limitations}).

Training hyperparameters and optimization details are provided in the Appendix~\pageref{app:appendix}.

\section{Experiments and Results}
\label{sec:comparative_experiments}

\subsection{Experimental Setup}

We evaluated \textit{TALL} on the task of predicting a missing Hebrew word at the end of a given sentence, comparing it against several baseline approaches:

\begin{itemize}
    \item \textbf{Direct Hebrew:} The LLM attempts to predict the missing Hebrew word directly.
    
    \item \textbf{Fine-tuned:} The LLM is fine-tuned on the news dataset.
    
    \item \textbf{From-Scratch:} A language model with identical architecture to the original LLM but trained entirely from scratch on the news dataset.
    
    \item \textbf{Naive:} The Hebrew sentence is translated to English, the LLM predicts the next English token, and the completed English sentence is translated back to Hebrew.
    
    \item \textbf{Soft Prompt:} A set of trainable soft prompt parameters is added to guide the LLM toward better predictions \cite{lester2021power}.
    
    \item \textbf{\textit{TALL}:} Our proposed architecture that integrates translation models with the LLM through alignment adapters.
\end{itemize}

We used two LLMs with different levels of Hebrew exposure: bloomz-560m \cite{muennighoff2022crosslingual} with minimal Hebrew exposure, and QWEN2.5-0.5b \cite{qwen2.5} with moderate Hebrew exposure. The evaluation was conducted on two datasets of 50,000 Hebrew sentences each: WIKI-SLVM from the Hebrew Wikipedia Corpus \cite{svlm2021hebrew} and HE-LYRICS sampled from the Hebrew Stage and Lyrics dataset \cite{norod2023hebrew}.

\subsection{Results and Analysis}
\label{sec:results_analysis}

Table~\ref{tab:approach_accuracy_comparison} presents the accuracy results for predicting the missing Hebrew word. \textit{TALL} consistently outperforms all baseline approaches, achieving up to 5.59\% accuracy with bloomz-560m and 5.15\% with QWEN2.5-0.5b on the WIKI-SLVM dataset, compared to the next best methods at 2.93\% and 3.99\% respectively.

Direct Hebrew prediction performs poorly (0.63\% with bloomz-560m), and surprisingly, fine-tuning often degrades performance further (0.16\%). Translation-based approaches show modest improvements over direct Hebrew prediction, suggesting that leveraging high-resource representations provides some benefit. However, these approaches fail to fully bridge the gap between languages. QWEN2.5-0.5b demonstrates higher baseline accuracy than bloomz-560m, reflecting its greater Hebrew exposure during pretraining, yet the relative improvement from \textit{TALL} is more pronounced with bloomz-560m.

Despite freezing approximately 86\% of parameters, \textit{TALL} achieves substantial improvements by effectively aligning representations between languages while preserving the strengths of pretrained models. This demonstrates that our architecture successfully leverages high-resource language capabilities to enhance performance in low-resource contexts without extensive retraining.

\section{Conclusion and Future Work}

This paper introduced \textit{TALL}, a novel architecture for enhancing LLM performance in low-resource languages. By integrating high-resource language LLMs with bilingual translation models through trainable alignment adapters, \textit{TALL} effectively leverages the strengths of pre-trained models while minimizing computational overhead.

Our experimental results on Hebrew demonstrate that \textit{TALL} significantly outperforms direct use of LLMs, naive translation approaches, models trained from scratch, and fine-tuning techniques. The architecture maintains strong performance across different domains, highlighting its robustness and generalization capabilities.

Key advantages of \textit{TALL} include:

\begin{itemize}
    \item \textbf{Parameter Efficiency:} By freezing large pre-trained components and training only lightweight adapters, \textit{TALL} achieves significant performance gains with minimal computational overhead.
    
    \item \textbf{Cross-domain Generalization:} \textit{TALL} demonstrates robust performance consistency across different text domains, maintaining its effectiveness when applied to diverse content types.
    
    \item \textbf{Modularity and Extensibility:} The architecture can be adapted to various low-resource languages by substituting the appropriate translation models, offering a flexible solution for multilingual NLP.
\end{itemize}

Future work will extend \textit{TALL} to additional low-resource languages and further refine the alignment mechanisms. By continuing to develop this approach, we aim to contribute to more inclusive and accessible NLP technologies that support all languages, regardless of their resource availability.

\section{Limitations} 
\label{sec:limitations}

While \textit{TALL} presents a promising framework for enhancing LLM performance in low-resource languages, several limitations warrant consideration:

\paragraph{Inference Efficiency Challenges}
The integration of multiple translation and alignment modules complicates efficient inference, particularly for key-value caching typically used in LLM pipelines. This results in slower inference compared to standard LLMs. Future work could explore partial caching techniques or optimized implementations to address this limitation.

\paragraph{Reliance on Translation Models}
\textit{TALL} depends on pre-trained translation models for the target language pair, which may introduce biases or errors inherent in these models. For extremely low-resource languages where quality translation models are unavailable, this dependency could limit applicability. Alternative approaches using multilingual translation models or zero-shot cross-lingual transfer could potentially address this challenge.

\paragraph{Single-Token Learning Signal}
Our current implementation focuses on predicting only the final token of a sequence, which limits the learning signal compared to conventional LLM training that predicts every token. This design choice effectively addresses our evaluation task but may constrain the model's ability to learn broader contextual dependencies. Future iterations could explore curriculum learning strategies or multi-stage training protocols with additional per-token predictions.

\paragraph{Error Propagation}
As a multi-stage pipeline, errors can accumulate and propagate across translation, alignment, and decoding stages. 

\paragraph{Adapter Layer Information Bottleneck}
The dimension alignment adapters, while necessary for integration, could introduce information bottlenecks that limit the maximum achievable performance. Future refinements may explore more expressive adapter architectures or non-linear alignment techniques to mitigate this limitation.

These limitations highlight clear research directions for improving the efficiency, generalizability, and performance of cross-lingual transfer approaches like \textit{TALL}.

\bibliography{custom}
\clearpage
\appendix

\section*{Appendix}
\label{app:appendix}

\subsection*{Model Training Details}

\subsubsection*{Fine-Tuning Configuration}
Our implementation utilized parameter-efficient fine-tuning with the following configuration:
\begin{itemize}
    \item \textbf{Optimizer:} AdamW ($\eta = 2 \times 10^{-5}$, weight decay = 0.01)
    \item \textbf{Training:} Max sequence length = 128, gradient clipping = 1.0
    \item \textbf{Model-specific optimizations:} Customized padding token handling for BLOOMZ, and Qwen architectures
\end{itemize}

\subsubsection*{From-Scratch Training}
For training models without pre-trained weights:
\begin{itemize}
    \item \textbf{Optimizer:} AdamW ($\eta = 5 \times 10^{-4}$, weight decay = 0.01)
    \item \textbf{Training strategy:} Higher learning rate, increased gradient accumulation (8 steps)
    \item \textbf{Efficiency:} Mixed precision (FP16) with dynamic loss scaling
\end{itemize}

\subsubsection*{Soft Prompt Training}
The soft prompt approach consisted of:
\begin{itemize}
    \item \textbf{Architecture:} Trainable prompt embeddings ($n = 30$) with frozen LLM parameters
    \item \textbf{Optimization:} AdamW ($\eta = 5 \times 10^{-4}$) with cosine schedule and 100 warmup steps
    \item \textbf{Implementation:} Input embeddings concatenation ($E_{prompt} \oplus E_{input}$) with extended attention masks
\end{itemize}

\subsubsection*{TALL Training Process}

The training of our TALL architecture employed specific techniques for efficient cross-lingual learning:

\begin{itemize}
    \item \textbf{Token-Focused Loss:} Training focused exclusively on predicting the final token of target sequences, with a specialized loss function that isolates the last non-padded position in each sequence
    
    \item \textbf{Optimization:} AdamW optimizer with cosine annealing schedule ($\eta$ decaying over epochs)
    
    \item \textbf{Evaluation Metrics:} Tracked per-token accuracy, loss, and perplexity with focused evaluation on final token prediction
    
    \item \textbf{Checkpoint Management:} Maintained best-model checkpoints based on evaluation loss, with gradient norm monitoring for training stability
\end{itemize}

\subsection*{Evaluation Framework}
Our comparative evaluation used a unified framework for all models:
\begin{itemize}
\item \textbf{Task:} Single-token Hebrew word prediction at sentence endings
\item \textbf{Approaches:} Direct Hebrew, Naive Translation, From-scratch, Fine-tuned, Soft Prompt, and TALL implementations
\item \textbf{Inference:} Temperature-controlled sampling ($T = 0.7$) with top-k (50) and top-p (0.95) filtering
\item \textbf{Metrics:} Per-token accuracy with progressive tracking across dataset samples
\end{itemize}

Each training approach prioritized parameter efficiency by optimizing only the minimal necessary subset of parameters.
All evaluation results, along with the full evaluator code, are available in our repository.

\begin{table*}[!t]
\centering
\begin{tabular}{lcc}
\hline
\textbf{Metric} & \textbf{bloomz TALL} & \textbf{Qwen TALL} \\
\hline
Total Parameters       & 883,537,920              & 799,239,680              \\
LLM Only Parameters    & 559,214,592 (63.3\%)       & 494,032,768 (61.8\%)       \\
Trainable Parameters   & 126,786,048 (14.35\%)      & 107,669,632 (13.47\%)      \\
\hline
\end{tabular}
\caption{Overall Model Statistics for bloomz TALL and Qwen TALL. The ``LLM Only Parameters'' row represents the sum of LLM Embeddings and Main LLM parameters.}
\label{tab:overall_stats}
\end{table*}

\begin{table*}[!t]
\centering
\begin{tabular}{llcc}
\hline
\textbf{Module}       & \textbf{Total Params} & \textbf{Trainable Params} & \textbf{Notes} \\
\hline
HE-EN Encoder         & 138,341,376  & 0           & Frozen encoder \\
LLM Embeddings        & 256,901,120  & 0           & Frozen embedding layer \\
Autoencoder 1         & 4,203,520    & 4,203,520   & Two-layer MLP (1024 $\rightarrow$ 2048, 2048 $\rightarrow$ 1024) \\
Custom Decoder 1      & 101,828,608  & 101,828,608 & Trainable decoder module \\
Main LLM              & 302,313,472  & 0           & Frozen main LLM (BloomModel) \\
Autoencoder 2         & 1,577,472    & 1,577,472   & Two-layer MLP (1024 $\rightarrow$ 1024, 1024 $\rightarrow$ 512) \\
Custom Encoder 2      & 19,176,448   & 19,176,448  & Trainable encoder module \\
EN-HE Decoder         & 59,195,904   & 0           & Frozen decoder module \\
LM Head               & 33,709,568   & 0           & Final linear mapping \\
\hline
\end{tabular}
\caption{Module-wise Breakdown for bloomz TALL.}
\label{tab:bloomz_breakdown}
\end{table*}

\begin{table*}[!t]
\centering
\begin{tabular}{llcc}
\hline
\textbf{Module}       & \textbf{Total Params} & \textbf{Trainable Params} & \textbf{Notes} \\
\hline
HE-EN Encoder         & 138,341,376  & 0           & Frozen encoder \\
LLM Embeddings        & 136,134,656  & 0           & Frozen embedding layer \\
Autoencoder 1         & 3,448,704    & 3,448,704   & Two-layer MLP (1024 $\rightarrow$ 1792, 1792 $\rightarrow$ 896) \\
Custom Decoder 1      & 83,598,080   & 83,598,080  & Trainable decoder module \\
Main LLM              & 357,898,112  & 0           & Frozen main LLM (Qwen2Model) \\
Autoencoder 2         & 1,446,400    & 1,446,400   & Two-layer MLP (896 $\rightarrow$ 1024, 1024 $\rightarrow$ 512) \\
Custom Encoder 2      & 19,176,448   & 19,176,448  & Trainable encoder module \\
EN-HE Decoder         & 59,195,904   & 0           & Frozen decoder module \\
LM Head               & 33,709,568   & 0           & Final linear mapping \\
\hline
\end{tabular}
\caption{Module-wise Breakdown for Qwen TALL.}
\label{tab:qwen_breakdown}
\end{table*}

\end{document}